\documentclass{article}
\usepackage{microtype}
\usepackage{graphicx}
\usepackage{subfigure}
\usepackage{booktabs} 
\usepackage{color}
\usepackage{amssymb}
\usepackage{amsmath}
\usepackage{multirow}
\usepackage [latin1]{inputenc}
\usepackage{hyperref}

\usepackage[accepted]{icml2021}

\begin{document}

\twocolumn[
\icmltitle{GCN-SL: Graph Convolutional Networks with Structure Learning for Graphs under Heterophily}



\icmlsetsymbol{equal}{*}

\begin{icmlauthorlist}
\icmlauthor{Mengying Jiang}{to}
\icmlauthor{Guizhong Liu}{to}
\icmlauthor{Yuanchao Su}{ed}
\icmlauthor{Xinliang Wu}{to}

\end{icmlauthorlist}
\icmlaffiliation{to}{School of Electronics and Information Engineering,Xi'an Jiaotong University, Xi'an 710049, China.}
\icmlaffiliation{ed}{College of Geomatics, Xi'an University of Science and Technology, Xi'an, 710054, China}
\icmlcorrespondingauthor{Mengying Jiang}{myjiang@stu.xjtu.edu.cn}
\icmlcorrespondingauthor{Guizhong Liu}{liugz@xjtu.edu.cn}
\icmlcorrespondingauthor{Yuanchao Su}{suych3@xust.edu.cn}
\icmlcorrespondingauthor{Xinliang Wu}{wuliang@stu.xjtu.edu.cn}

\vskip 0.3in
]



\printAffiliationsAndNotice{}  

\begin{abstract}
In representation learning on the graph-structured data, under heterophily (or low homophily), many popular GNNs may fail to capture long-range dependencies, which leads to their performance degradation. To solve the above-mentioned issue, we propose a graph convolutional networks with structure learning (GCN-SL), and furthermore, the proposed approach can be applied to node classification. The proposed GCN-SL contains two improvements: corresponding to node features and edges, respectively. In the aspect of node features, we propose an efficient-spectral-clustering (ESC) and an ESC with anchors (ESC-ANCH) algorithms to efficiently aggregate feature representations from all similar nodes. In the aspect of edges, we build a re-connected adjacency matrix by using a special data preprocessing technique and similarity learning. Meanwhile, the re-connected adjacency matrix can be optimized directly along with GCN-SL parameters. Considering that the original adjacency matrix may provide misleading information for aggregation in GCN, especially the graphs being with a low level of homophily. The proposed GCN-SL can aggregate feature representations from nearby nodes via re-connected adjacency matrix and is applied to graphs with various levels of homophily. Experimental results on a wide range of benchmark datasets illustrate that the proposed GCN-SL outperforms the state-of-the-art GNN counterparts.
\end{abstract}

\section{Introduction}
Graph Neural Networks (GNNs) are powerful for representation learning on graphs with varieties of applications ranging from knowledge graphs to financial networks~\citep{{Gilmer+chapter2017},{BGNN+chapter2020},{gdl+chapter2017}}. In recent years, GNNs have developed many artificial neural networks, e.g.,
Graph Convolutional Network (GCN)~\citep{GCN+chapter2017}, Graph Attention Network (GAT)~\citep{GAT+chapter2018}, Graph Isomorphism Network (GIN)~\citep{GIN+chapter2019}, and GraphSAGE~\citep{GraphSage+chapter2017}.
For GNNs, each node can iteratively update its feature representation via aggregating the ones of the node itself and its neighbors~\citep{GCN+chapter2017}.
The neighbors are usually defined as all adjacent nodes in graph, meanwhile a diversity of aggregation functions can be adopted to GNNs~\citep{{Geom+chapter2020},{pna+chapter2020}}, e.g.,
summation, maximum, and mean.

GCNs is one of the most attractive GNNs and have been widely applied in a variety of scenarios~\citep{{Geom+chapter2020},{hamilton+chapter2017},{gdl+chapter2017}}.
However, one fundamental weakness of GCNs limits the representation ability of GCNs on graph-structured data, i.e., GCNs may not capture long-range dependencies in graphs, considering that GCNs update the feature representations of nodes via simply summing the normalized feature representations from all one-hop neighbors~\citep{{GAT+chapter2018},{Geom+chapter2020}}.
Furthermore, this weakness will be magnified in graphs with
heterophily or low/medium levels of homophily.

Homophily is a very important principle of many real-world graphs, whereby the linked nodes tend to have similar features and belong to the same class \citep{hhgcn}.
For instance, papers are more likely to cite papers from the same research area in citation networks, and friends tend to have similar age or political beliefs in social networks.
However, there are also settings about "opposites attract" in the real world, leading to graphs with low homogeneity, i.e., the proximal nodes are usually from different classes and have dissimilar features~\citep{hhgcn}.
For example, most of people tend to chat with people of the opposite gender in dating website, and fraudsters are more prefer to contact with accomplices than other fraudsters in online gambling networks.
Most existing GNNs assume strong homophily, including GCNs, therefore they perform poorly on generalizing graphs under high heterophily even worse than the MLP~\citep{MLP} that relies only on the node features for classification~\citep{fraudsters,hhgcn}.

For addressing the mentioned weakness,
recently some related approaches has been bulit to alleviating this curse, such as Geometric Graph Convolutional Network (GEOM-GCN)~\citep{Geom+chapter2020}, H$_{2}$GCN~\citep{hhgcn}.
Although GEOM-GCN improves the performance of representation learning of GCNs, the performance of node classification is often unsatisfactory if the concerned datasets is the graphs with low homogeneity~\citep{Geom+chapter2020}.
H$_{2}$GCN improves the classification performance of GCN, while it is only able to aggregate information from near nodes, resulting in lacking an ability for capturing the features from distant but similar nodes.
Nevertheless, it is notable that H$_{2}$GCN still has a lot of room for improvement.

In this paper, we propose a novel GNN approach to solve the above mentioned problem, referred to as the graph convolutional networks with structure learning (GCN-SL).
Following the spectral clustering (SC) method~\citep{sc+chapter2011},
a graph of data points is built according to the distances of the data points in the feature space.
Then, the data points are mapped into a new feature space by cutting the graph. Furthermore, the data points connected closely in graph are usually proximal in new feature space. Therefore, nodes can aggregate features from similar nodes if SC is employed to process graph-structured data, contributing to that GCN can capture long-range dependencies. It should be noted, however, the computational complexity of SC is greatly high for large-scale graphs. Hence, we design an efficient-spectral-clustering(ESC)
and an ESC with anchors (ESC-ANCH) algorithms to efficiently extract SC features.
Then, the extracted SC features combined with the original node features as enhanced features (EF), and the EF is utilized to train the proposed GCN-SL.

Following the research in~\citet{hhgcn}, the nodes of the same class always possess similar features, no matter whether the homophily of graph is high or low.
For our approach, it can build a re-connected graph related to the similarities between nodes, and the re-connected graph is optimized jointly with the GCN-SL parameters.
Meanwhile, for dealing with sparse initial attributes of nodes and overcoming the problem of over-fitting that GNNs with structure learning module faced,
a special data preprocessing technique is applied to the initial features of nodes.
Afterwards, the original adjacency matrix and the re-connected adjacency matrix are utilized respectively to obtain multiple intermediate representations corresponding to different rounds of aggregation. Then, we combine several key intermediate representations as the new node embedding, and use a learnable weighted vector to highlight the important dimensions of the new node embedding.
Finally, we set the result of this calculation as the final node embedding for node classification.

Compared with other GNNs, the contributions of GCN-SL can be summarized as follows. 1) SC is integrated into GNNs for capturing long range dependencies on graphs, and an ESC and an ESC-ANCH algorithms are proposed to efficiently implement SC on graph-structured data; 2) Our GCN-SL can learn an optimized  re-connected adjacency matrix which benefit the downstream prediction task. Meanwhile, the special data preprocessing technique used not only overcomes the problem of over-fitting that most of GNNs with structure learning module faced but also benefits the generation of the re-connected adjacency matrix; 3) The GCN-SL proposes the improvements for handling heterophily from the aspects of node features and edges, respectively. Meanwhile, GCN-SL combines the two improvements and makes them supplement each other.

\section{Related Work}
\label{related work}

In this paper, our work is directly related to SC and GCN. The theories about the SC and GCN are reviewed in subsection~\ref{Spectral Clustering} and subsection~\ref{GCN}, respectively.

\subsection{Spectral Clustering}
\label{Spectral Clustering}

Spectral clustering (SC) is an advanced algorithm evolved from graph theory, has attracted much attention~\citep{sc+chapter2011}. Compared with most traditional clustering methods, the implementation of SC is much simpler~\citep{sc+chapter2011}. It is remarkable that, for SC, a weighted graph is utilized to partition the dataset.
Assume that $\mathcal{X}=\left\{{\bf x}_{i}\right\}_{i=1}^{n}$ represents a dataset.
The task of clustering is to segment $\mathcal{X}$ into $c$ clusters.
The cluster assignment matrix is denoted as
${\bf F}=\left[{\bf f}_{1}, {\bf f}_{2}, \ldots, {\bf f}_{n}\right]^{T} \in  \mathbb{R}^{n \times c}$,
where $\mathbf{f}_{i}\in \mathbb{R}^{c \times 1}(1 \leq i \leq n)$ is the cluster assignment vector for the pattern $\mathbf{x}_{i}$.
From another perspective, $\mathbf{f}_{i}$ can be considered as the feature representation of $\mathbf{x}_{i}$ in the $c$-dimensional feature space.
SC contains different types, while our work focuses on the SC with $k$-way normalized cut due to considering the overall performance of the algorithm,
where the related concepts are explained in~\citet{sc-ncut}.

Let $\mathcal{G}=\{\mathcal{X}, \bf S\}$ be an undirected weighted graph,
where $\mathcal{X}$ denotes the set of nodes, and ${\bf S} \in \mathbb{R}^{n \times n}$ denotes affinity matrix, and $n$ is the number of nodes in $\mathcal{G}$.
Note that ${\bf S}$ is symmetric matrix, and each element $S_{i, j}$ represents the affinity of a pair of nodes in $\mathcal{G}$.
The most common way to built ${\bf S}$ is full connection method.
Following the description in~\citet{sc+chapter2011}, $S_{i, j}$ can be represented as:
\begin{equation}
S_{i j}=\exp \left(-\frac{\left\|{\bf x}_{i}-{\bf x}_{j}\right\|^{2}}{2 \sigma^{2}}\right),~~                         i,j =1,2, \ldots, n
\label{eq:1}
\end{equation}
where $\mathbf{x}_{i}$ and $\mathbf{x}_{j}$ represent the features of $i$-th node and $j$-th node in $\mathcal{X}$, respectively,
and $\sigma$ can control the degree of similarity between nodes.
Then,
the Laplacian matrix $\bf L$ is defined as $\bf L=\bf D-\bf S$, where degree matrix $\bf D$ is a diagonal matrix, and the diagonal element is denoted as $D_{i, i}=\sum_{j} S_{i, j}, \forall i$.

The objective function of SC with normalized cut is
\begin{equation}
\min _{{\bf F}^{T} {\bf F}={\bf I}}tr \left({\bf F}^{T} {\bf D}^{-1 / 2} {\bf L} {\bf D}^{-1 / 2} {\bf F}\right)
\label{eq:2}
\end{equation}
where ${\bf F} \in \mathbb{R}^{n \times c}$ is the clustering indicator matrix, and the objective function can be rewritten as:
\begin{equation}
\max _{{\bf F}^{T} {\bf F}={\bf I}} tr \left({\bf F}^{T} {\bf D}^{-1 / 2} {\bf S} {\bf D}^{-1 / 2} {\bf F}\right)
\label{eq:3}
\end{equation}

The optimal solution ${\bf F}$ of objective function is constructed by the eigenvectors corresponding to the $c$ largest eigenvalues of ${\bf G}={\bf D}^{-1 / 2} {\bf S} {\bf D}^{-1 / 2}$.
In general, ${\bf F}$ can be not only considered as the result of clustering of nodes, but also regarded as the new feature matrix of nodes, in which node has $c$ feature elements.
\begin{figure*}[t]
\centering
\vspace{-0.2cm}
\includegraphics[width=6.7in,height=2.3in]{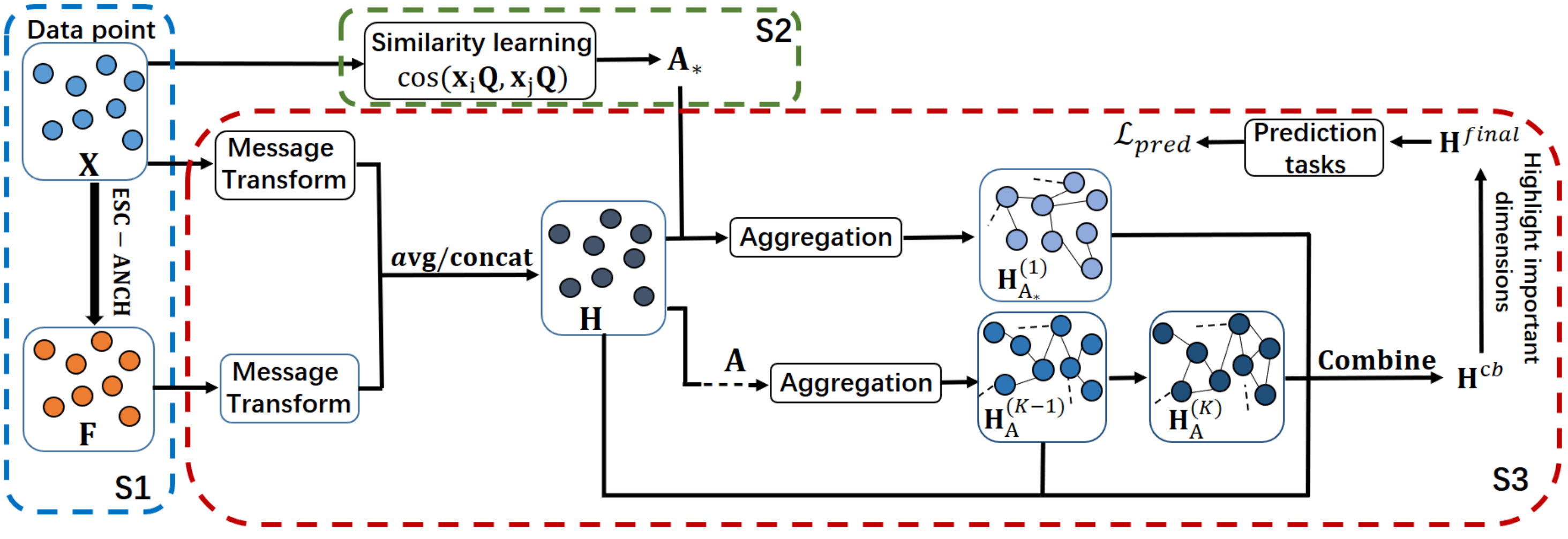}
\hspace{-0.3cm}
\caption{GCN-SL pipeline. It consists of three stages:
(S1) \textit{efficient spectral clustering with anchors}, (S2) \textit{re-connected graph}, and (S3) \textit{structure learning graph convolutional networks}.
In (S1), we employ ESC-ANCH to generate the SC feature ${\bf F}$.
In (S2), we construct a re-connected graph according to the similarities between nodes, and the re-connected graph can be gradually optimized with the training of the GCN-SL model.
In (S3), we combine the original feature ${\bf X}$ and SC feature ${\bf F}$ as the enhanced feature ${\bf H}$, and perform feature aggregation on ${\bf H}$ by using the re-connected adjacency matrix ${\bf A}_{*}$ and original ${\bf A}$, respectively.
Then, the results of aggregation and the ${\bf H}$ are combined as the ${\bf H}^{cb}$,
and a weighted vector is used to highlight important dimensions of ${\bf H}^{cb}$ so as to make GCN-SL adapts to graphs with various levels of homophily.
}
\label{fig:Fig.1}
\end{figure*}
\subsection{GCN}

\label{GCN}
Following the work in~\citet{{BGNN+chapter2020}}, in GCN, the updates of node features adopt an isotropic averaging operation over the feature representations of one-hop neighbors.
Let ${\bf h}_{j}^{\ell}$ be the feature representation of node $j$ in the $l$-th GCN layer, and we have
\begin{equation}
{\bf h}_{i}^{\ell+1} =  {\rm ReLU}\left(\sum_{j \in \mathcal{N}_{i}} \frac{1}{\sqrt{{\rm deg}_{i}{\rm deg}_{j}}} {\bf h}_{j}^{\ell} {\bf W}^{\ell}\right)
\label{eq:GCNlayer}
\end{equation}

where $\mathcal{N}_{i}$ represents the set of one-hop neighbors of nodes $i$, ${\bf W}^{\ell}$ is a learnable weight matrix, and ${\rm ReLU}$ is employed as the activation function. Note that ${\rm deg}_{i}$ and ${\rm deg}_{j}$ are the in-degrees of nodes $i$ and $j$, respectively.
Furthermore, the forward model of a 2-layer GCN can be represented as:
\begin{equation}
{\bf Z}=\displaystyle f({\bf X}, {\bf A})=\displaystyle softmax\left(\hat{{\bf A}} {\rm ReLU}\left(\hat{{\bf A}} {\bf X} {\bf W}^{0}\right) {\bf W}^{1}\right)
\label{eq:2layerGCN}
\end{equation}
where ${\bf X}\in \mathbb{R}^{n \times d}$ is the feature matrix of nodes and is also the input of first GCN layer, and
${\bf A}\in \mathbb{R}^{n \times n}$ is the adjacency matrix.
The adjacency matrix with self-loop is $\tilde{{\bf A}}={\bf A}+{\bf I}$, where ${\bf I}$ is an identity matrix.
Here, the adjacency matrix with self-loop can be normalized by $\hat{{\bf A}}=\widetilde{{\bf D}}^{-1 / 2} \tilde{{\bf A}} \widetilde{{\bf D}}^{-1 / 2}$,
and the normalized adjacency matrix $\hat{{\bf A}}$ is employed for aggregating the feature representation of neighbors, where $\tilde{{\bf D}}$ is the degree matrix of $\tilde{{\bf A}}$. Each element $\hat{A}_{i, j}$ is defined as:

\begin{equation}
\hat{A}_{i, j}=\left\{\begin{array}{cl}
\frac{1}{\sqrt{{\rm deg}_{i}{\rm deg}_{j}}} &{\rm nodes~} i, j {\rm ~are~one-hop~neighbors} \\
0 & {\rm otherwise }
\end{array}\right.
\label{eq:Aij}
\end{equation}

\section{Proposed GCN-SL Approach}
\label{headings}

In this section, we present a novel GNN: graph convolutional networks with structure learning (GCN-SL) for node classification of graph-structured data. The pipeline of GCN-SL is shown in Figure~\ref{fig:Fig.1}.
The remainder of this section is organized as follows.
Subsection~\ref{FSC} describes the proposed ESC and ESC-ANCH approaches.
In subsection~\ref{adjacent}, we give the generated details of the re-connected adjacency matrix.  Subsection~\ref{gcn-ef} describes the proposed GCN-SL in detail.

\subsection{Efficient Spectral Clustering with Anchors}\label{FSC}
In graph-structured data, a large number of nodes of the same class possess similar feature but are far apart from each other.
However, GCN simply aggregates the information from one-hop neighbors, and the depth of GCN is usually limited.
Therefore, the information from distant but similar nodes is always ignored, while SC can divide nodes according to the affinities between nodes.
Specifically, the closely connected and similar nodes are more likely to be proximal in new feature space, and vice versa.
Thus, it is very appropriate for combining GCN with SC to extract the features of distant but similar nodes.
Following subsection~\ref{Spectral Clustering}, the object of performing SC is to generate the cluster assignment matrix ${\bf F}$.
${\bf F}$ can only be calculated by eigenvalue decomposition on the normalized similar matrix ${\bf G}= {\bf D}^{-1 / 2} {\bf S} {\bf D}^{-1 / 2}$, which takes $O(n^2 c)$ time complexity, where $n$ and $c$ are the the numbers of nodes and clusters, respectively.
For some large-scale graphs, the computational complexity is an unbearable burden.

In order to overcome this problem, we propose the efficient-spectral-clustering (ESC) to efficiently perform SC.
In the proposed ESC, instead of the S is calculated by equation~(\ref{eq:1}),
we employ inner product to construct affinity matrix ${\bf S}={\bf X}{\bf X}^T$.
Thus, the normalized similar matrix ${\bf G}$ in the ESC method can be represented as:

\begin{equation}
\begin{aligned}
{\bf G} &= {\bf D}^{-1 / 2}{\bf X}{\bf X}^{T} {\bf D}^{-1 / 2}\\
&= ({\bf D}^{-1 / 2}{\bf X})({\bf D}^{-1 / 2}{\bf X})^T
\label{eq:K}
\end{aligned}
\end{equation}

Then, we define ${\bf P} = {\bf D}^{-1 / 2}{\bf X}$, thus ${\bf G}={\bf P} {\bf P}^T$.
The singular value decomposition (SVD) of ${\bf P}$ can be represent as follows:
\begin{equation}
{\bf P} = {\bf U} \Sigma {\bf V}^{T}
\label{eq:SVD}
\end{equation}
where ${\bf U}\in \mathbb{R}^{n \times n}$, $\Sigma\in \mathbb{R}^{n \times d}$ and ${\bf V}\in \mathbb{R}^{d \times d}$ are left singular vector matrix, singular value matrix and right singular vector matrix, respectively.
It is obvious that the column vectors of ${\bf U}$ are the eigenvectors of ${\bf G}$.
Therefore, we can easily construct ${\bf F}$ by using the eigenvectors from ${\bf U}$, and the eigenvectors used are
corresponding to the $c$ largest eigenvalues in $\Sigma$.
The computational complexity of SVD performed on ${\bf P}\in \mathbb{R}^{n \times d}$ ($O=ndc$) is much lower, compared with directly performing eigenvalue decomposition on ${\bf G}\in \mathbb{R}^{n \times n}$ ($O=nnc$).

However, the dimension $d$ of nodes' original features are usually high in many graph-structured data, the efficiency of the ESC method still needs to be improved.

Therefore, we propose the ESC-ANCH.
Specifically, we first randomly select $m$ nodes from the set of nodes $\mathcal{X}$ as the anchor nodes, where $m<<n$ and $m<d$.
Then, we calculate the node-anchor similar matrix ${\bf R}\in \mathbb{R}^{n \times m}$, and the chosen similarity metric function is cosine.
Here, ${\bf R}$ is employed as the new feature representations of nodes.
Thus, in ESC-ANCH, ${\bf P}$ can be redefined as ${\bf P}={\bf D}^{-1 / 2}{\bf R}$, and {\bf D} is the degree matrix of ${\bf S}={\bf R}{\bf R}^T$.
Therefore, the ESC-ANCH only takes $O(nmc)$ time complexity to perform the singular value decomposition of ${\bf P}={\bf D}^{-1 / 2}{\bf R}$, which performs more efficient than SC and ESC.
In this paper, we default to use ESC-ANCH to obtain SC features.
\subsection{Re-connected Graph}\label{adjacent}

\begin{figure}[t]\vspace{-3mm}
\centering
\vspace{2mm}
\includegraphics[width=3.2in,height=1.1in]{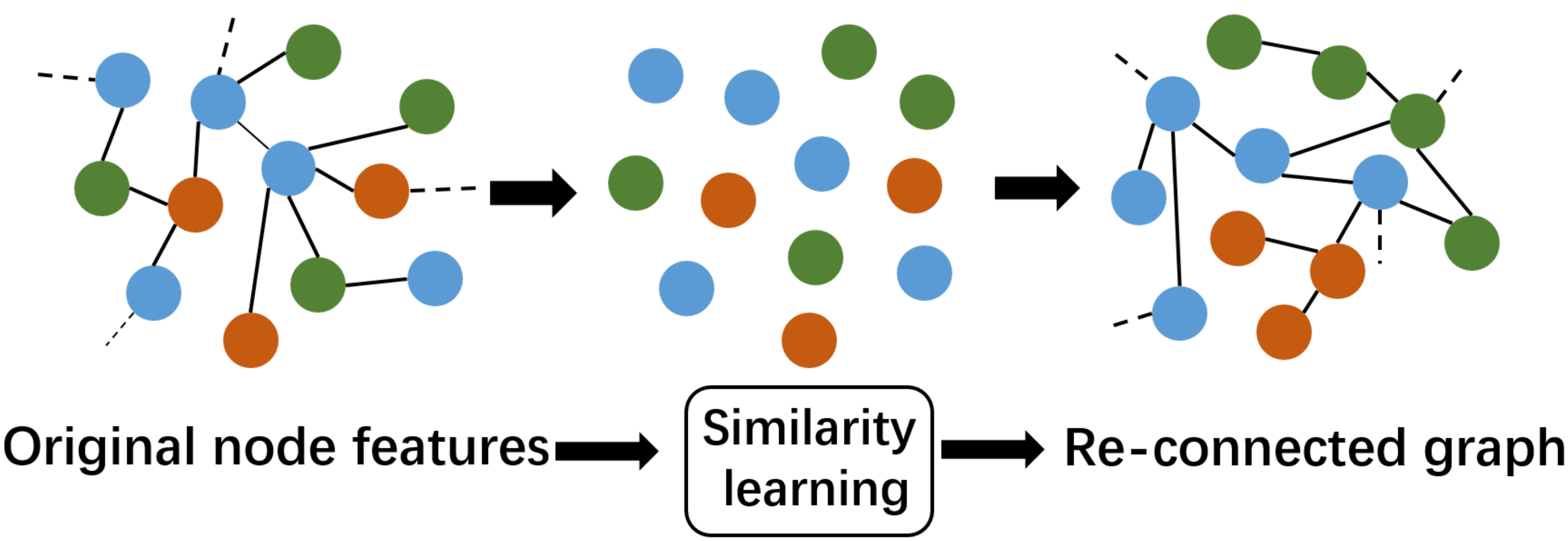}
\caption{Reconnected graph is learned from original node features via similarity learning.}
\vspace{-2mm}
\label{similar}  
\end{figure}

Most existing GNNs are designed for graphs with high homophily, where the linked nodes are more likely possess similar feature representation and belong to the same class, e.g., community networks and citation networks~\citep{amn}.

However, there are a large number of real-world graphs under heterophily, where the linked nodes usually possess dissimilar feature and belong to different classes, e.g., webpage linking networks~\citep{struc2vec}.
Thus, these GNNs designed under the assumption of homophily are greatly inappropriate for the graphs under heterophily.

Fortunately, regardless of the homophily level of graphs, the nodes of the same class always possess high similar feature.
Therefore, as shown in Figure~\ref{similar}, for helping GCN to capture information from nodes in the same class, we learn a re-connected adjacency matrix ${\bf A}_{*}\in \mathbb{R}^{n \times n}$ according to the similarities between nodes and the downstream tasks.

A good similarity metric function should be expressively powerful and learnable~\citep{IDGL}.
Here, we set cosine similarity as our metric function.
The similarity between a pair of nodes can be represented as:
\begin{equation}
S_{i j}=\cos \left(\mathbf{x}_{i}{\bf Q} ,\mathbf{x}_{j}{\bf Q} \right)
\label{eq:cosine}
\end{equation}
where $\mathbf{x}_{i}$ and $\mathbf{x}_{j}$ represent the features of $i$-th node and $j$-th node, and ${\bf Q}\in \mathbb{R}^{d \times p}$ is a learnable weighted vector.
Afterwards, we can generate the similarity matrix ${\bf S}$.
The generated similarity matrix {\bf S} is symmetric, and the elements in {\bf S} range between $[-1,1]$.

However, an adjacency matrix is supposed to be non-negative and sparse.
This is owning to the fully connected graph is very computational and might introduce noise.
Therefore, we need to extract a non-negative and sparse adjacency matrix ${\bf A}_{*}$ from ${\bf S}$.
Specifically, we define a non-negative threshold $\epsilon$, and set
those elements in ${\bf S}$ which are smaller than $\epsilon$ to zero.

Since graph structure learning module enable the GCN with stronger ability to fit the downstream task, it would be easier for them to over-fitting.
Meanwhile, the sparse initial attributes of nodes usually result in an invalid similarity matrix ${\bf S}$.
In order to overcome the mentioned weakness, we apply a special data preprocessing technique to the ${\bf X}$.
Specifically, we randomly choose a feature dimension from ${\bf X}$, and plus a number greater than $0$ to the chosen feature dimension.
The introduction of additional features help GCN-SL to overcome the over-fitting problem.
Meanwhile, in order to guarantee the calculation of ${\bf S}$, and not affect similarity of nodes too much, the number greater than zero is set as 0.5 for all graph-structured data.

\subsection{Structure Learning Graph Convolutional Networks}
\label{gcn-ef}
\begin{algorithm}[tb]
   \caption{GCN-SL}
   \label{alg:example}
\begin{algorithmic}
   \STATE {\bfseries Input:} ${\bf X}$: feature matrix of nodes, ${\bf Y}$: label matrix of nodes for training,
   ${\bf A}$: original adjacency matrix, ${\bf Q}, {\bf W}^{0}_{X}, {\bf W}^{0}_{F}, {\bf W}^{1}$: trainable weighted matrixs, ${\bf w}$: trainable weighted vector.
   \STATE {\bfseries Output:} ${\bf Z}$: probability distribution of nodes over a set of classes.
   \STATE 1. ${\bf F}\leftarrow{\bf X}$ in~\ref{FSC}.
   \STATE 2. ${\bf X}\leftarrow \left \{{\bf X}, Data~preprocessing\right \}$.
   \STATE 3. ${\bf A}_{*}\leftarrow \left \{  {\bf X}, {\bf Q}\right \}$ in~\ref{adjacent}.
   \STATE 4. ${\bf H}\leftarrow \left \{  {\bf X}, {\bf F}, {\bf W}^{0}_{X},{\bf W}^{0}_{F}\right \}$
   in the way of average in Eq.~\ref{eq:GCN-EF-AGG}, or in the way of concatenation in Eq.~\ref{eq:GCN-EF-CAT}.
   \STATE 5. ${\bf H}^{(1)}_{A_{*}}\leftarrow \left \{ {\bf H}, {\bf A}_{*}\right \}$ in Eq.~\ref{eq:2-GCN-EF1}.
   \STATE {\bfseries repeat} (initialize $k$ to 1)
   \STATE 6. Update ${\bf H}^{(k)}_{A}\leftarrow \left \{ {\bf H}^{(k-1)}_{A},{\bf A}\right \}$, ${\bf H}^{(0)}_{A}={\bf H}$ in Eq.~\ref{eq:2-GCN-EF}, $k=k+1$.
   \STATE {\bfseries until $k=K$}
   \STATE 7. ${\bf H}^{cb} \leftarrow \left \{{\bf H},{\bf H}^{(K-1)}_{A},{\bf H}^{(K)}_{A}, {\bf H}^{(1)}_{A_{*}}\right \}$ in Eq.~\ref{eq:2-GCN-EF2}.
   \STATE 8. ${\bf H}^{final} \leftarrow \left \{{\bf H}^{cb}, {\bf w}\right \}$ in Eq.~\ref{hadamard}
   \STATE 9. ${\bf Z} \leftarrow \left \{{\bf H}^{final}, {\bf W}^{1}\right \}$ in Eq.~\ref{eq:2-GCN-EF3}.
   \STATE 10. $\mathcal{L}_{pred}\leftarrow $ LOSS $({\bf Z}, {\bf Y})$ in Eq.~\ref{eq:2-GCN-EF4}.
   \STATE 11. Back-propagate $\mathcal{L}_{pred}$ to update model weights.

\end{algorithmic}
\end{algorithm}

We first utilize original feature ${\bf X}$ and SC feature ${\bf F}$ to construct the EF.
The first layer of the proposed GCN-SL is represented as:
\begin{equation}
{\bf H}= \text{ReLU}\left( \frac{{\bf X} {\bf W}^{0}_{X}+{\bf F} {\bf W}^{0}_{F}}{2} \right)
\label{eq:GCN-EF-AGG}
\end{equation}
where ${\bf W}^{0}_{X}\in \mathbb{R}^{d \times q}$, ${\bf W}^{0}_{F}\in \mathbb{R}^{c \times q}$ are trainable weight matrix of the first layer.
Meanwhile, we can also use concatenation to combine ${\bf X}$ and ${\bf F}$.
\begin{equation}
{\bf H} = \text{ReLU}\left({\bf X} {\bf W}^{0}_{X}\| {\bf F} {\bf W}^{0}_{F} \right)
\label{eq:GCN-EF-CAT}
\end{equation}
where $\|$ represents concatenation.
Here, we call the EF as EF$_{av}$ if average methd is adopted, and call EF as EF$_{cc}$ if concatenation method is adopted.

After the first layer is constructed,
We use ${\bf A}_{*}$ generated from subsection~\ref{adjacent} to aggregate and update features of nodes to obtain the intermediate representations:
\begin{equation}
{\bf H}^{(1)}_{A_{*}} = \hat{{\bf A_{*}}}{\bf H}
\label{eq:2-GCN-EF1}
\end{equation}
where $\hat{{\bf A_{*}}}$ is the row normalized ${\bf A_{*}}$.
Similarly, we use ${\bf A}$ to aggregate and update features to obtain the intermediate representations ${\bf H}^{(k)}_{A}$ of nodes.
\begin{equation}
{\bf H}^{(k)}_{A} = \hat{{\bf A}}{\bf H}^{(k-1)}_{A}
\label{eq:2-GCN-EF}
\end{equation}
where $k=1,2,\ldots,K$, and
$K$ represents the times of feature aggregation.
${\bf H}^{(0)}_{A} = {\bf H}$.

After several rounds of feature aggregation, we combine several most key intermediate representations as the new node embeddings as follows:
\begin{equation}
{\bf H}^{cb} = \operatorname{COMBINE}({\bf H},
{\bf H}^{(K-1)}_{A},{\bf H}^{(K)}_{A}, {\bf H}^{(1)}_{A_{*}})
\label{eq:2-GCN-EF2}
\end{equation}

\label{experiment}\vspace{-3mm}
\begin{table*}[t]
\caption{Summary of the datasets utilized in our experiments.}
\label{Tab:des-datasets}
\begin{center}
\begin{tabular}{l c c c c c c c c }
\toprule
Dataset& Cora & Citeseer & Pubmed & Squirrel & Chameleon & Cornell & Texas & Wisconsin \\ \hline  
\ Hom.ratio $h$   & 0.81 & 0.74  & 0.8   &0.22&0.23&0.3 &0.11&0.21\\
\# Nodes           & 2708 & 3327 & 19717 &5201&2277&183 &183 &251\\
\# Edges           & 5429 & 4732 & 44338 &198493&31421&295 &309 &499\\
\# Features        & 1433 & 3703 & 500   &2089&2325&1703&1703&1703\\
\# Classes         & 7    & 6    & 3     &5&5& 5  &5   &5  \\
\bottomrule
\end{tabular}
\end{center}
\end{table*}

For graphs with high homophily, ${\bf H}^{(K-1)}_{A}$ and ${\bf H}^{(K)}_{A}$ are sufficient for representing embeddings of nodes. This can be proved by GCN~\citep{{GCN+chapter2017}} and GAT~\citep{{GAT+chapter2018}}.
In addition, ${\bf H}^{(1)}_{A_{*}}$ can be treated as the supplement to ${\bf H}^{(K-1)}_{A}$ and ${\bf H}^{(K)}_{A}$.
For graphs with under heterophily, ${\bf H}$ and ${\bf H}^{(1)}_{A_{*}}$ can also perform well on learning feature representation.
In order to exploit the advantage of these intermediate representations to the full, we use concatenation to combine these intermediate representations. The intermediate representations adopted are not mixed, by the way of concatenation.

Afterwards, we generate a learnable weight vector ${\bf w}$ which has the same dimension as ${\bf H}^{cb}$,
and take the Hadamard product between ${\bf w}$ and ${\bf H}^{cb}$ as the final feature representations of nodes,
\begin{equation}
{\bf H}^{final} = \text{ReLU}({\bf w} \odot {\bf H}^{cb})
\label{hadamard}
\end{equation}
The purpose of this step is highlight the important dimensions of ${\bf H}^{cb}$.
Afterward, nodes are classified based on their final embedding ${\bf H}^{final}$ as follows:
\begin{equation}
{\bf Z} =\displaystyle softmax\left({\bf H}^{final}{\bf W}^{1}\right)
\label{eq:2-GCN-EF3}
\end{equation}
where ${\bf W}^{1}$ is trainable weighted matrix of the last layer.
The classification loss is as follows:
\begin{equation}
\mathcal{L}_{\text {pred}} =\ell({\bf Z}, {\bf Y})
\label{eq:2-GCN-EF4}
\end{equation}
where {\bf Y} represent the labels of nodes, and $\ell(\cdot, \cdot)$ is cross-entropy function for measuring the difference between predictions ${\bf Z}$ and the true labels ${\bf Y}$.

In this paper, the GCN-SL that uses EF$_{cc}$ is called as GCN-SL$_{cc}$, and similarly, the GCN-SL that uses EF$_{av}$ is named as GCN-SL$_{av}$.
A pseudocode of the proposed GCN-SL is given in Algorithm~\ref{alg:example}.

\section{Experimental Results}

Here, we validate the merit of GCN-SL by comparing GCN-SL with some state-of-the-art GNNs on transductive node classification tasks on a variety of open graph datasets.
\subsection{Datasets}

In simulations, we adopt three common citation networks, three sub-networks of the WebKB networks, and two Wikipedia networks to validate the proposed GCN-SL.

Citation networks, i.e., Cora, Citeseer, and Pubmed are standard citation network benchmark datasets~\citep{ccnd}, ~\citep{Query-driven}.
In these datasets, nodes correspond to papers and edges correspond to citations. Node features represent the bag-of-words representation of the paper, and the label of each node is the academic topics of the paper.

Sub-networks of the WebKB networks , i.e., Cornell, Texas, and Wisconsin.
They are collected from various universities' computer science departments~\citep{{Geom+chapter2020}}.
In these datasets, nodes correspond to webpages, and edges represent hyperlinks between webpages. Node features denote the bag-of-words representation of webpages.
These webpages can be divided into 5 classes.

Wikipedia networks, i.e., Chameleon and Squirrel are page-page networks about specific topics in wikipedia. Nodes correspond to pages, and edges correspond to the mutual links between pages.
Node features represent some informative nouns in Wikipedia pages.
The nodes are classified into four categories based on the amount of their average traffic.

For all datasets, we randomly split nodes per class into $48\%$, $32\%$, and $20\%$ for
training, validation and testing, and the experimental results are the mean and standard deviation of ten runs. Testing is performed when validation losses achieves minimum on each run.
An overview summary of characteristics of all datasets are shown in Table~\ref{Tab:des-datasets}.

The level of homophily of graphs is one of the most important characteristics of graphs,
it is significant for us to analyze and employ graphs.
Here, we utilize the edge homophily ratio $h=\frac{\left|\left\{(u, v):(u, v) \in \mathcal{E} \wedge y_{u}=y_{v}\right\}\right|}{|\mathcal{E}|}$ to describe the homophily level of a graph, where $\mathcal{E}$ donate the set of edges,
$y_{v}$ and $y_{u}$ respectively represent the label of node $v$ and $u$.
The $h$ is the fraction of edges in a graph which linked nodes that have the same class label (i.e., intra-class edges).
This definition is proposed in~\citet{hhgcn}.
Obviously, graphs have strong homophily when $h$ is high ($h \rightarrow 1$), and graphs have strong heterophily or weak homophily when $h$ is low ($h \rightarrow 0$).
The $h$ of each graph are listed in Table~\ref{Tab:des-datasets}.
From the homophily ratios of all adopted graph, we can find that all the citation networks are high homogeneous graphs, and all the WebKB networks and wikipedia networks are low homogeneous graphs.
\begin{figure*}[th]\vspace{-3mm}
\centering
\includegraphics[width=6.6in,height=1.7in]{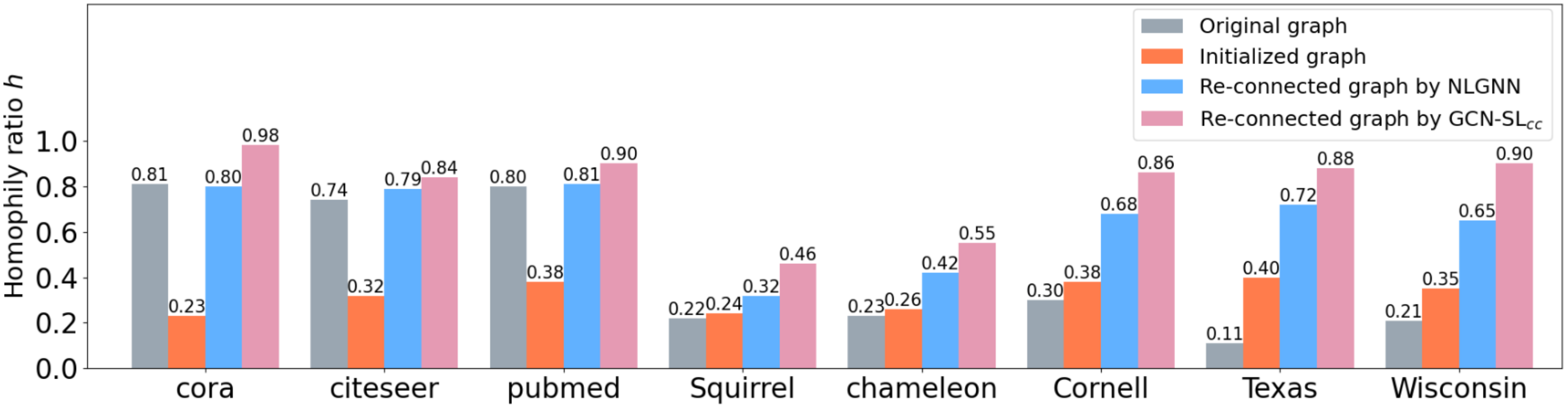}
\caption{Comparisons of homophily between the original graph, initialized graph and the re-connected graph on networks.}
\vspace{-5mm}
\label{homo}  
\end{figure*}
\begin{table*}\vspace{3mm}
\caption{Node Classification Accuracies (\%) of the Proposed GCN-SL$_{cc}$ and GCN-SL$_{av}$ for All Graphs}
\label{re-connected}
\begin{center}
\begin{tabular}{l c c c c c c c c c c c }
\toprule
Dataset&${\bf A}$& ${\bf A}_{*}$ &Cora & Citeseer & Pubmed &Squirrel&Chamele.&Cornell & Texas & Wiscons. \\ \hline
             &--&--&74.7$\pm$1.9&72.1$\pm$1.8&87.1$\pm$1.4&29.4$\pm$2.9&49.4$\pm$2.9&82.6$\pm$5.5&82.3$\pm$3.9&86.5$\pm$2.1\\
GCN-SL$_{cc}$&\checkmark&--&\bf{89.1$\pm$0.4}&\bf{77.3$\pm$1.5}&88.5$\pm$1.1&37.1$\pm$3.2&57.5$\pm$2.8&76.6$\pm$8.6&81.2$\pm$6.2&81.8$\pm$5.3\\
             &\checkmark&\checkmark&88.8$\pm$1.3&77.3$\pm$1.5&89.7$\pm$1.2&\bf{38.9$\pm$3.1}&\bf{60$\pm$3.2}&\bf{85.9$\pm$5.1}&\bf{85.6$\pm$6.2}&86.6$\pm$4.8\\ \hline
             &--&--&75.2$\pm$2.7&71.7$\pm$2.7&86.9$\pm$0.6&30.2$\pm$1.4&48.7$\pm$3.1&82.4$\pm$5.9&82.1$\pm$4.8&86.2$\pm$2.2\\
GCN-SL$_{av}$&\checkmark&--&88.7$\pm$0.6&77$\pm$1.5&\bf{89.9$\pm$0.8}&35.6$\pm$3.0&55.9$\pm$3.7&76.8$\pm$5.9&81.9$\pm$3.2&84$\pm$6.0\\
             &\checkmark&\checkmark&\bf{89.1$\pm$1.1}&77$\pm$2.1&89.5$\pm$0.8&38.6$\pm$2.2&59.1$\pm$1.2&85.3$\pm$6.3&85$\pm$5.9&\bf{86$\pm$4.2}\\
\bottomrule
\end{tabular}
\end{center}
\end{table*}
\subsection{Baselines}
We compare our proposed GCN-SL$_{av}$ and GCN-SL$_{cc}$ with various baselines:

$\bullet$ MLP~\citep{MLP} is the simplest deep neural networks. It just makes prediction based on the feature vectors of nodes, without considering any local
or non-local information.

$\bullet$ GCN~\citep{GCN+chapter2017} is the most common GNN. As introduced in Section~\ref{GCN}, GCN makes prediction solely aggregating local information.

$\bullet$ GAT~\citep{GAT+chapter2018} enable specifying different weights to different nodes in a neighborhood by employing the attention mechanism.

$\bullet$ MixHop~\citep{MixHop} model can learn relationships by repeatedly mixing feature representations of neighbors at various distances.

$\bullet$ Geom-GCN~\citep{Geom+chapter2020} is recently proposed and can capture long-range dependencies.
Geom-GCN employs three embedding methods, Isomap, struc2vec, and Poincare embedding, which make for three Geom-GCN variants: Geom-GCN-I, Geom-GCN-S,
and Geom-GCN-P. In this paper, we report the best results of three Geom-GCN variants without specifying the employed embedding method.

$\bullet$ H$_{2}$GCN~\citep{hhgcn} identifies a set of key designs: ego and
neighbor embedding separation, higher order neighborhoods.
H$_{2}$GCN adapts to both heterophily and homophily by effectively synthetizing these designs.
We consider two variants: H$_{2}$GCN-1 that uses one embedding round (K = 1) and H$_{2}$GCN-2 that uses two rounds (K = 2).

\begin{figure*}[t]\vspace{5mm}
\centering
\subfigure[\hspace{-0.4cm}]{ \label{fig:subfig:4a} 
\includegraphics[width=2.22in,height=1.58in]{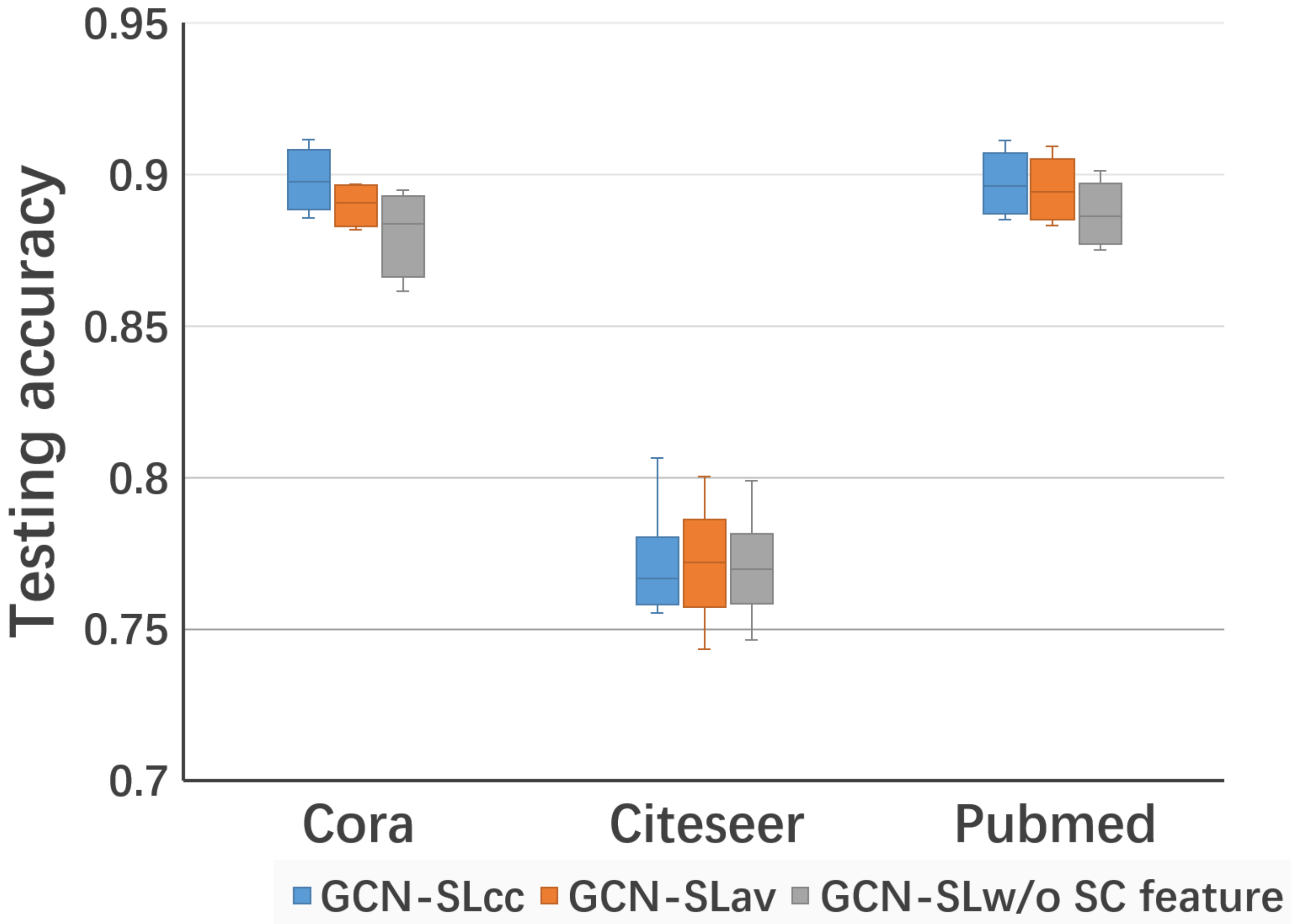}}
\hspace{-0.3cm}
\subfigure[\hspace{-0.4cm}]{ \label{fig:subfig:4b} 
\includegraphics[width=2.22in,height=1.58in]{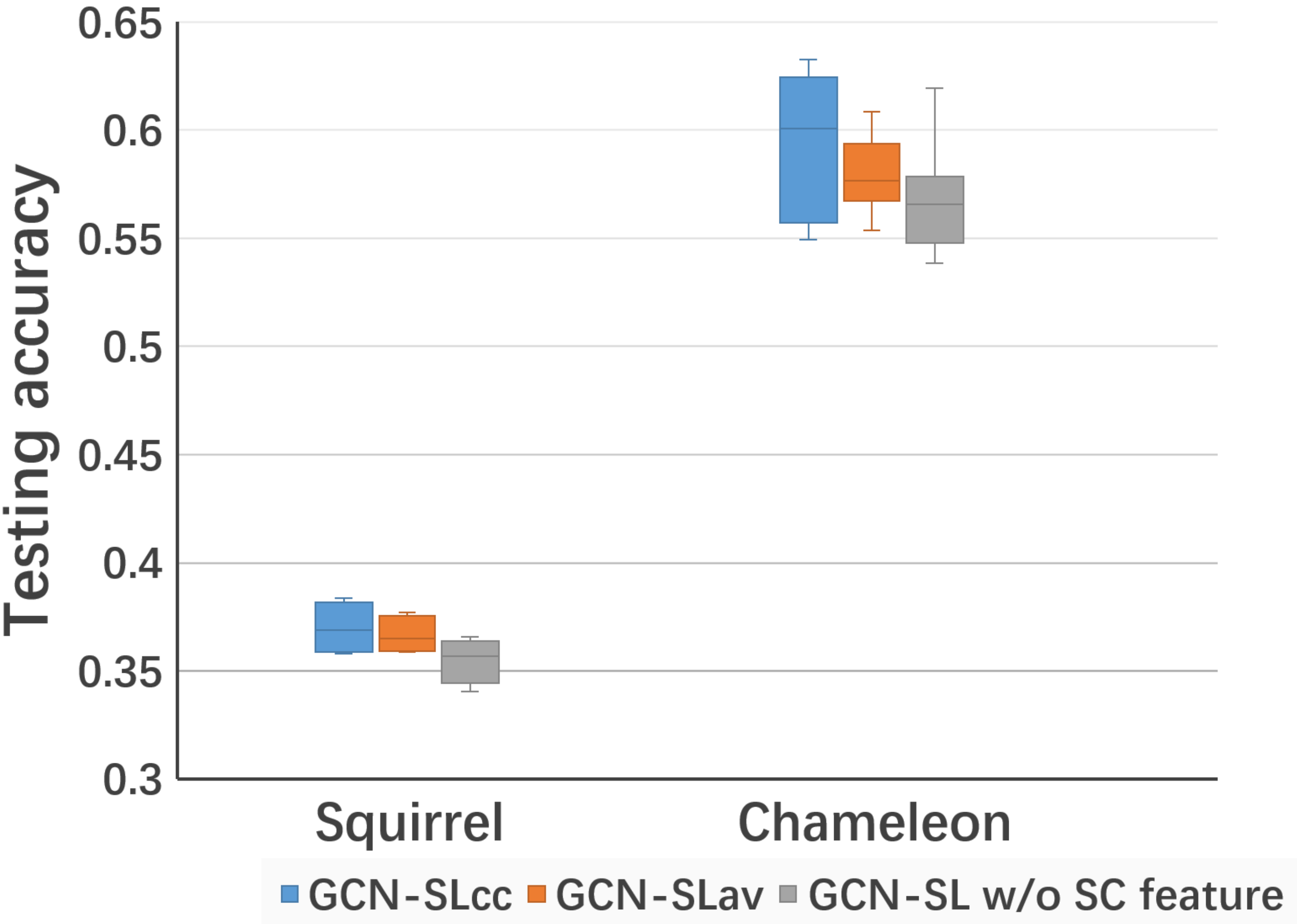}}
\hspace{-0.3cm}
\subfigure[\hspace{-0.4cm}]{ \label{fig:subfig:4c} 
\includegraphics[width=2.22in,height=1.58in]{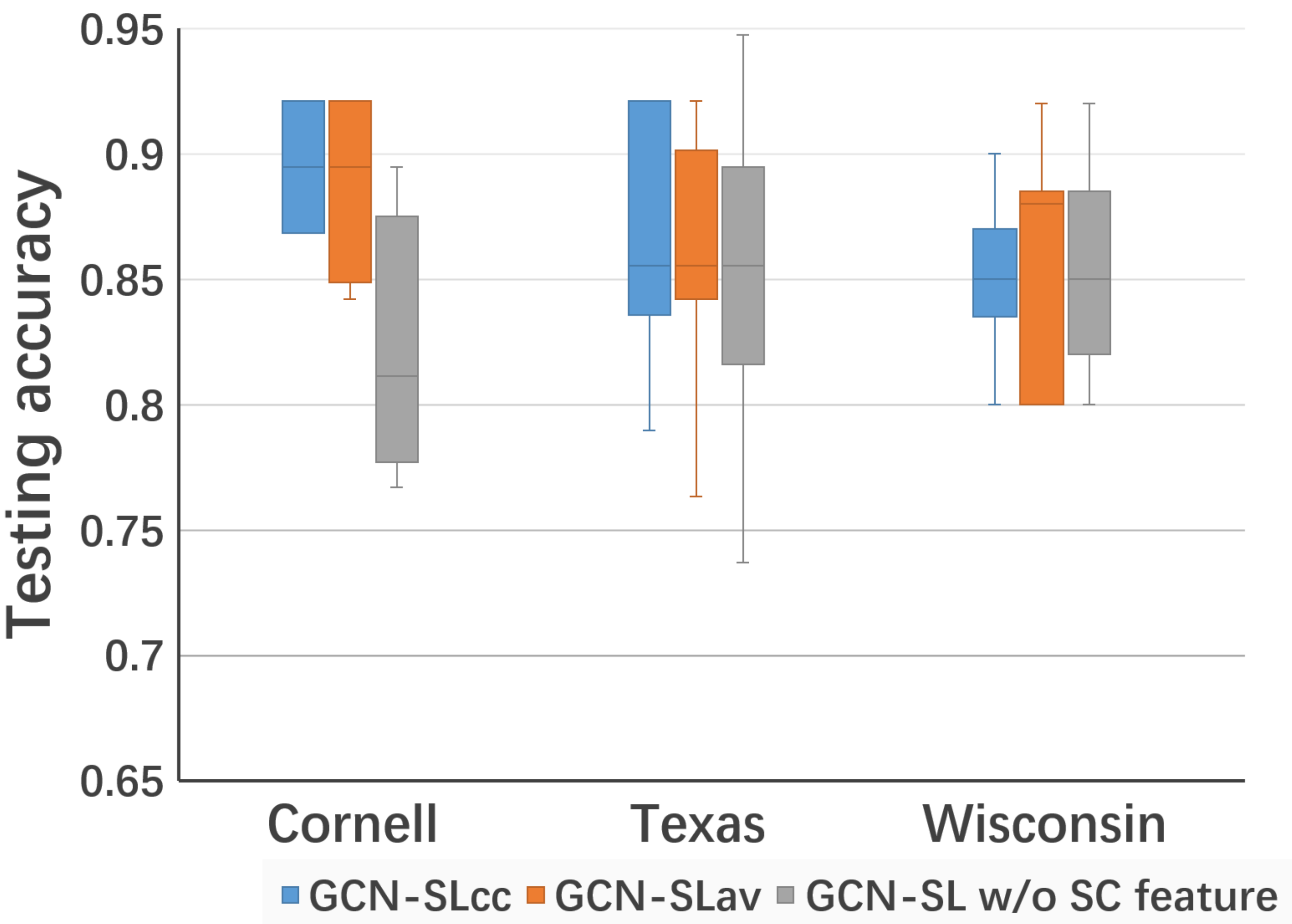}}
 \vspace{-0.2cm}
\caption{Ablation study about SC feature on the proposed GCN-SL.}\vspace{-2mm}
\label{scccc}  
\end{figure*}
\begin{table*}[t]\vspace{-2mm}
\caption{Running Time (Seconds) of SC, the Proposed ESC and ESC-ANCH on All Graph Datasets (OM Error)}
\label{Tab:running}
\begin{center}
\begin{tabular}{l c c c c c c c c }
\toprule
Method & Cora & Citeseer & Pubmed &  Squirrel &Chameleon & Cornell & Texas & Wisconsin \\ \hline  
SC  & 58.1 & 63.2 & OM    &98.5&29.3&1.7 &1.7 &1.8\\
ESC & 1.7  & 4.6 & 2.0     &2.5& 2.2&1.8 &1.8 &1.9\\
ESC-ANCH & \bf{1.2}  & \bf{0.98}  & \bf{1.4} &\bf{1.1}& \bf{0.96}&\bf{0.96} &\bf{0.83} &\bf{0.89} \\
\bottomrule
\end{tabular}
\end{center}
\end{table*}
\subsection{Experimental Setup}

For comparison,
we use six state-of-the-art node classification algorithms, i.e.,
MLP, GCN, GAT, MixHop, GEOM-GCN, and H$_{2}$GCN.
For the above six GNNs, all hyper-parameters are set according to~\citet{hhgcn}.
For the proposed GCN-SL, we perform a hyper-parameter search on validation set.
The description of hyper-parameters that need to be searched are provided in Table~\ref{hyperparameters}.
Table~\ref{hyperparameters_value} summaries the hyper-parameters and accuracy of GCN-SL with the best accuracy on different datasets.
We utilize RELU as activation function. All models are trained to minimize
cross-entropy on the training nodes.
Adam optimizer is employed for all models ~\citep{{{adam}}}.

\subsection{Does Re-connected Adjacency Matrix Work?}

In this experiment, we first examine the correctness of the re-connected adjacency matrix ${\bf A}_{*}$ obtained via structure learning,
and the metric for judging the correctness of ${\bf A}_{*}$ is homophily ratios $h$. Figure~\ref{homo} shows the $h$ of original graph, initialized graph and re-connected graphs in all graphs.
The initialized graph is generated by GCN-SL before training, and
the re-connected graphs are learned by NLGNN and GCN-SL, respectively.
We can observe that the homophily ratios of re-connected graphs are much higher than the original graph and initialized graphs, especially for WebKB networks and Wikipedia networks.
This is mainly because of the re-connected graphs generated by either NLGNN or GCN-SL are constructed in terms of the similarities between nodes, and the weighted parameters involved in similarity learning can be optimized jointly with the parameters of the model. Thus, the re-connected graphs have a strong ability to fit the downstream task.
Meanwhile, the homophily ratios of the re-connected graphs GCN-SL learned are higher than the NLGNN learned.
This is owning to the similarity metric function chosen by NLGNN is inner product, which is not powerful enough to represent the similarities between nodes.
Meanwhile, the special data preprocessing technique utilized in GCN-SL can relieve the over-fitting problem, and is also greatly useful for structure learning.
In addition, compared with NLGNN, the ability to representation learning of the proposed GCN-SL is more capable, so GCN-SL can offer more help to re-connected graph learning.

Afterwards, we explore the impact of the learned re-connected graph on the accuracy of the proposed GCN-SL by using ablation study.
Table~\ref{re-connected} exhibits the accuracies of GCN-SL in all adopted networks.
We can see that the original adjacency matrix $\bf{A}$ is very important for GCN-SL$_{av}$ and GCN-SL$_{cc}$ in citation networks.
However, the impact of $\bf{A}$ is very limited and even bad in WebKB networks.
This is because citation networks are high homophily networks,
but WebKB networks are low homophily networks.
It is difficult for GCN-SL to aggregate useful information by only using $\bf{A}$ in graphs with low homophily.
By contrast, the introduction of re-connected adjacent matrix $\bf{A}_{*}$ is greatly helpful for accuracy of GCN-SL in WebKB networks, and does not hurt the performance of GCN-SL in citation networks.
Meanwhile, the adoption of $\bf{A}_{*}$ is also helpful for GCN-SL in Wikipedia networks.
This is owning to the re-connected graphs is constructed by similarity learning, and optimized as the model is optimized. Thus, the re-connected graphs are more reliable than original graph in low homophily networks.
Therefore, the introduction of re-connected graph can help GCN-SL$_{av}$ and GCN-SL$_{cc}$ adapt to all levels of homophily.

\subsection{Effect of SC Feature on Accuracy}

In this experiment, we explore the impact of SC feature extracted by the proposed ESC-ANCH method on the node classification accuracy of GCN-SL.
Here, we still use ablation study.
The classification accuracies of GCN-SL$_{cc}$, GCN-SL$_{av}$, and GCN-SL without SC feature are shown in Figure~\ref{scccc}, where all graph datasets are adopted.
As can be seen,
GCN-SL$_{cc}$ and GCN-SL$_{av}$ get better performance than GCN-SL without SC feature.
This owning to the the SC features not only reflect the ego-embedding of nodes but also the similar nodes.

As a further insight, we focus on the running times of original SC, the proposed ESC and ESC-ANCH in all adopted graph datasets.
From Table~\ref{Tab:running}, we can see that ESC and ESC-ANCH are much more efficient than SC.
Meanwhile, ESC-ANCH is faster than ESC due to the introduction of anchor nodes.
Specifically, ESC-ANCH only takes 1.2 s in Cora dataset, which is 47 times faster than original SC method.
Moreover, ESC-ANCH only takes 1.4 s, and original SC can not work in PubMed dataset due to "out-of-memory (OM) error."
In Squirrel networks, ESC-ANCH takes 1.1 s, which is 90 times faster than original SC method.
ESC-ANCH takes about half as long as original SC in WebKB networks.
In conclusion, ESC-ANCH is a very efficient SC method.

\begin{figure*}[t]\vspace{-3mm}
\centering
\subfigure[\hspace{-0.4cm}]{ \label{fig:subfig:4a} 
\includegraphics[width=2.25in,height=1.68in]{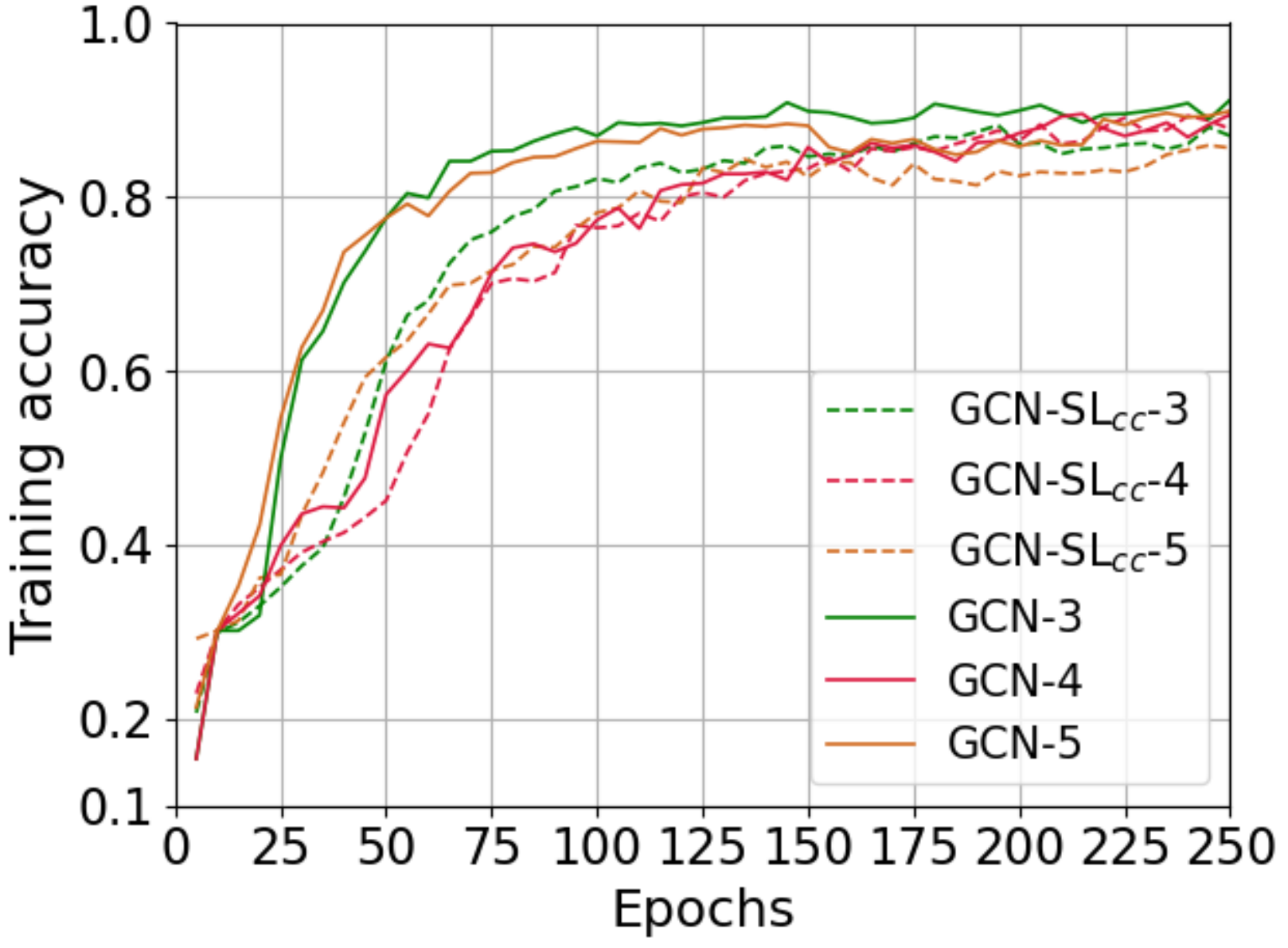}}
\hspace{-0.3cm}
\subfigure[\hspace{-0.4cm}]{ \label{fig:subfig:4b} 
\includegraphics[width=2.25in,height=1.68in]{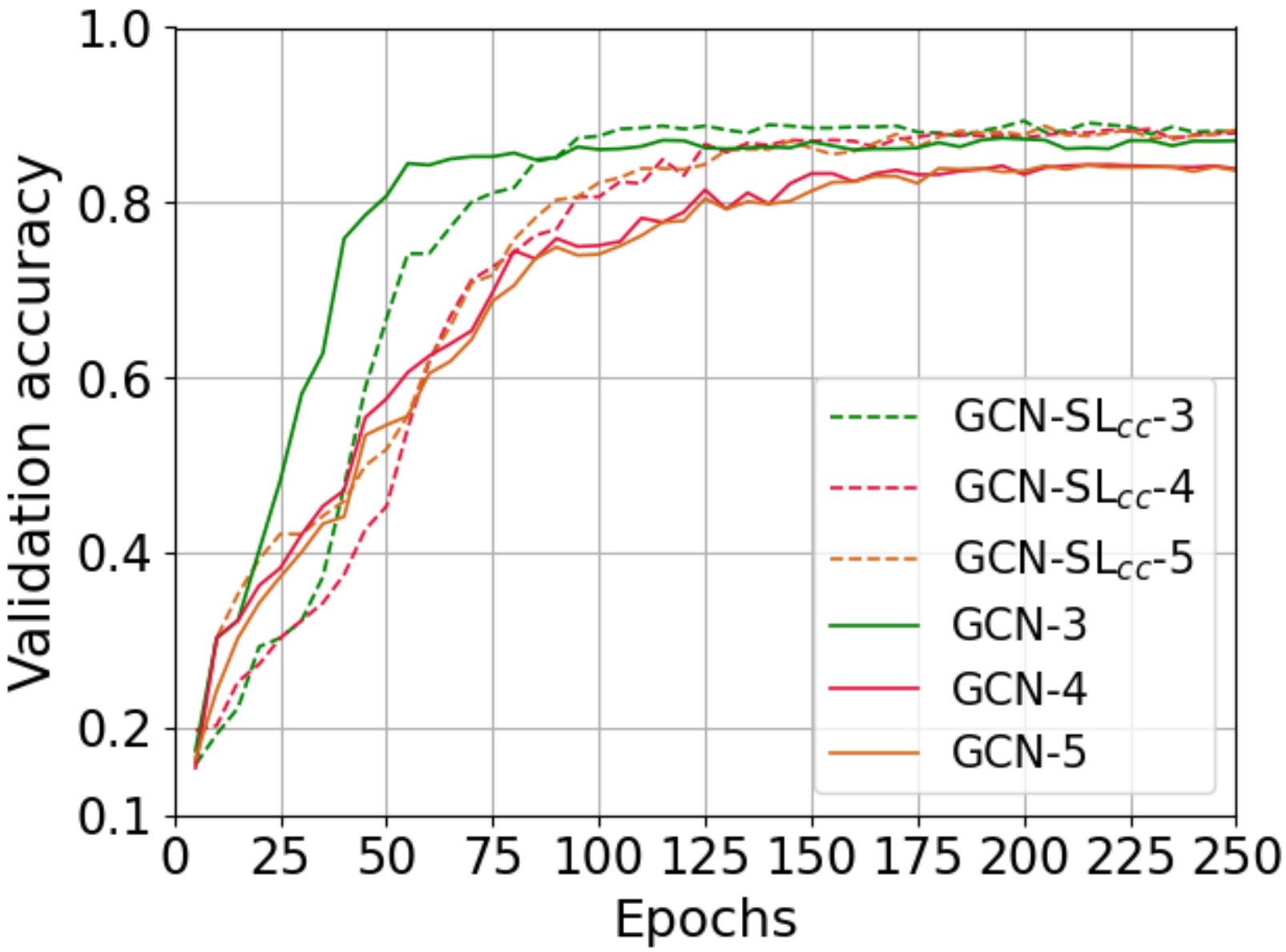}}
\hspace{-0.4cm}
\subfigure[\hspace{-0.4cm}]{ \label{fig:subfig:4c} 
\includegraphics[width=2.25in,height=1.68in]{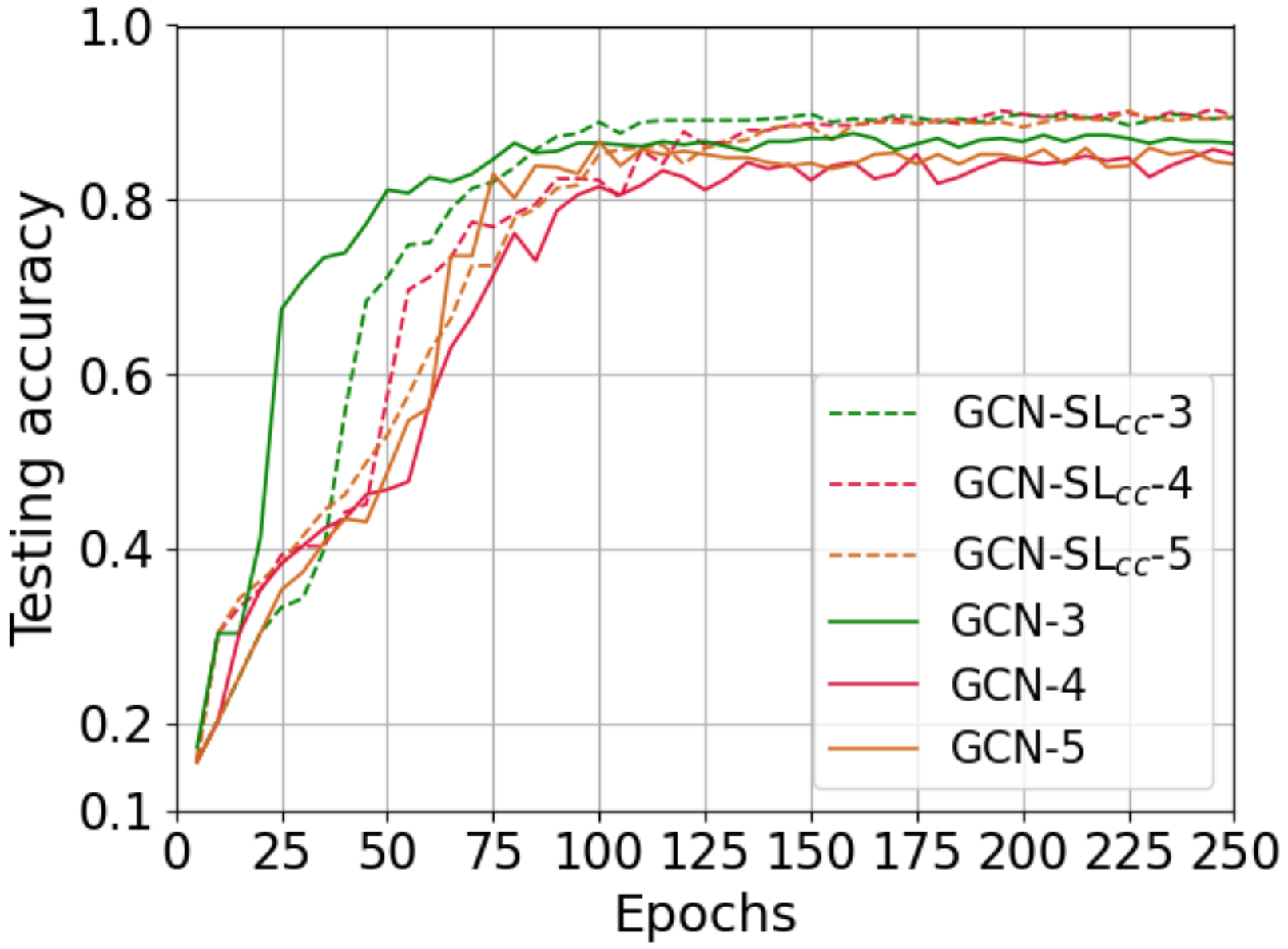}}
 \vspace{-0.4cm}
\caption{Performance of Multi-layer GCNs and the proposed GCN-SL$_{cc}$ on Cora. GCN-SL$_{cc}$-3, 4, 5 respectively represent the GCN-SL$_{cc}$ with different rounds of aggregation $K$, i.e., $K$ respectively equal to 3, 4, and 5.
}\vspace{-2mm}
\label{ACC}  
\end{figure*}
\begin{table*}[t]\vspace{-2mm}
\caption{Classification Accuracies (\%) of the MLP, GCN, GAT, Geom-GCN*, MixHop, H$_{2}$GCN and the Proposed GCN-SL for All Graphs.}
\label{all}
\begin{center}
\begin{tabular}{l c c c c c c c c }
\toprule
Method & Cora & Citeseer & Pubmed &  Squirrel &Chameleon & Cornell & Texas & Wisconsin \\ \hline  
MLP & 74.8$\pm$2.2 & 72.4$\pm$2.2 & 86.7$\pm$0.4&29.7$\pm$1.8&46.4$\pm$2.5&81.1$\pm$6.4&81.9$\pm$4.8 &85.3$\pm$3.6\\
GCN & 87.3$\pm$1.3&76.7$\pm$1.6&87.4$\pm$0.7&36.9$\pm$1.3&59.0$\pm$4.7&57.0$\pm$4.7&59.5$\pm$5.2 &59.8$\pm$7\\
GAT & 82.7$\pm$1.8&75.5$\pm$1.7&84.7$\pm$0.4&30.6$\pm$2.1&54.7$\pm$2.0&58.9$\pm$3.3&58.4$\pm$4.5 &55.3$\pm$8.7 \\
Geom-GCN* &85.3&78.0&\bf{90.1}&38.1&60.9&60.8&67.6&64.1\\
MixHop&87.6$\pm$0.9& 76.3$\pm$1.3 & 85.3$\pm$0.6 &\bf{43.8$\pm$1.5}& \bf{60.5$\pm$2.5}&73.5$\pm$6.3 &77.8$\pm$7.7&75.9$\pm$4.9  \\
H$_{2}$GCN-1&86.9$\pm$1.4&77.1$\pm$1.6&89.4$\pm$0.3&36.4$\pm$1.9&57.1$\pm$1.6&82.2$\pm$4.8&84.9$\pm$6.8&86.7$\pm$4.7 \\
H$_{2}$GCN-2& 87.8$\pm$1.4&76.9$\pm$1.7&89.6$\pm$0.3&37.9$\pm$2.0&59.4$\pm$2.0&82.2$\pm$6.0&82.2$\pm$5.3&85.9$\pm$4.2\\
GCN-SL$_{cc}$&\bf{89.2$\pm$1.3}&\bf{77.3$\pm$1.5}&89.7$\pm$1.2&38.9$\pm$3.1&60.0$\pm$3.2&\bf{85.9$\pm$5.1}&\bf{85.6$\pm$6.2}&\bf{86.6}$\pm$4.8\\
GCN-SL$_{av}$&89.1$\pm$1.1&77$\pm$2.1&89.5$\pm$0.8&38.6$\pm$2.2&59.1$\pm$1.2&85.3$\pm$6.3&85.0$\pm$5.9&86.1$\pm$4.2\\
\bottomrule
\end{tabular}
\end{center}
\end{table*}

\subsection{Comparison Among Different GNNs}
In Figure~\ref{ACC}, we implement 3-layer GCN, 4-layer GCN, 5-layer GCN and the proposed GCN-SL$_{cc}$ with different rounds of aggregation $K$ on Cora, i.e., $K$ respectively equal to 3, 4, and 5.
As can be seen from the validation and testing accuracies, the performance of multi-layer GCNs get worse as the depth increases.
This may be caused by over-fitting and over-smoothing.
While, GCN-SL$_{cc}$ perform well and do not suffer from over-fitting and over-smoothing as the depth increases.
Obviously, the round of aggregation in GCN-SL$_{cc}$-5 is $5$, so GCN-SL$_{cc}$-5 aggregate the same features of original neighbors as 5-layer GCN.
This shows that the reason for the performance degradation of GCN is not over-smoothing but over-fitting as the depth increases.
In addition, this also shows that GCN-SL can be immune to over-fitting no matter how many node features are aggregated.
This is because of the structural advantage, i.e., GCN-SL do not need to increase the depth of the network to aggregate more node features.

Table~\ref{all} gives classification results per benchmark.
We observe that all GNN models can achieve satisfactory result in citation networks.
However, some GNN models perform even worse than MLP in WebKB networks
and Wikipedia networks, i.e., GCN, GAT, Geom-GCN, and MixHop.
The main reason for this phenomenon is these GNN models aggregate useless information from wrong neighborhoods, and do not separate ego-embedding and the useless neighbor-embedding.
By contrast, H$_{2}$GCN and GCN-SL can achieve good classification results no matter which graphs adopted.
Meanwhile, the results of GCN-SL$_{cc}$ and GCN-SL$_{av}$ are relatively better than H$_{2}$GCN.
This mainly thanks to the re-connected graph of GCN-SL$_{cc}$ and GCN-SL$_{av}$.
In addition, the introduction of SC feature can improve the ego-embedding of nodes by clustering similar nodes together.

\begin{table}[t]
\caption{Hyper-parameter Description}
\label{hyperparameters}
\begin{center}
\begin{tabular}{c |c}
\toprule
Hyper-para.& Description\\ \hline  
$lr$  & learning rate  \\
$wd$ & weight decay \\
$d$ & dropout rate  \\
$\epsilon$ & non-negative threshold in similarity learning\\
$m$ & number of anchor nodes   \\
$c$ & dimension of SC features\\
$seed$ & random seed\\
\multirow{2}*{$p$} & output dimension of learnable weighted \\
& matrix ${\bf W}^{0}_{X}$ and ${\bf W}^{0}_{F}$\\
\multirow{2}*{$K$} & the rounds of aggregation based on original\\
&graph structure ${\bf A}$\\
\multirow{2}*{$q$} & output dimension of learnable weighted\\
& matrix ${\bf Q}$ \\

\bottomrule
\end{tabular}
\end{center}
\end{table}
\begin{table*}[t]\vspace{-2mm}
\caption{The hyper-parameters of best accuracy for GCN-SL on all datasets.}
\label{hyperparameters_value}
\begin{center}
\begin{tabular}{c|c |c}
\toprule
\multirow{2}*{Dataset}& Accuracy&\multirow{2}*{Hyper-parameters}\\
       &GCN-SL$_{cc}$~~~GCN-SL$_{av}$&  \\\hline  
Cora & 89.2$\pm$1.3~~~89.1$\pm$1.1&$lr$:0.02, $wd$:5e-4, $d$:0.5, $p$:32, $\epsilon$:0.9, $m$:700, $c$:75, $K$:4, $q$:16, $seed$ = 42\\
Citeseer &77.3$\pm$1.5~~~77.0$\pm$2.1&$lr$:0.02, $wd$:5e-4, $d$:0.5, $p$:32, $\epsilon$:0.98, $m$:500, $c$:30, $K$:3, $q$:16, $seed$ = 42\\
Pubmed &89.7$\pm$1.2~~~89.5$\pm$0.8&$lr$:0.02, $wd$:5e-4, $d$:0.5, $p$:32, $\epsilon$:0.95, $m$:500, $c$:25, $K$:2, $q$:16, $seed$ = 42\\

Squirrel &38.9$\pm$3.1~~~38.6$\pm$2.2 &$lr$:0.01, $wd$:5e-4, $d$:0.6, $p$:32, $\epsilon$:0.55, $m$:400, $c$:15, $K$:2, $q$:16, $seed$ = 42\\

Chameleon &60.0$\pm$3.2~~~59.1$\pm$1.2 &$lr$:0.005, $wd$:5e-4, $d$:0.6, $p$:32, $\epsilon$:0.95, $m$:200, $c$:13, $K$:3, $q$:16, $seed$ = 42\\

Cornell &85.9$\pm$5.1~~~85.3$\pm$6.3 &$lr$:0.01, $wd$:5e-4, $d$:0.4, $p$:48, $\epsilon$:0.55, $m$:100, $c$:15, $K$:1, $q$:16, $seed$ = 42\\
Texas &85.6$\pm$6.2~~~85.0$\pm$5.9 & $lr$:0.01, $wd$:5e-4, $d$:0.4, $p$:32, $\epsilon$:0.8, $m$:100, $c$:35, $K$:1, $q$:16, $seed$ = 42\\
Wisconsin & 86.6$\pm$4.8~~~86.1$\pm$4.2  &$lr$:0.01, $wd$:5e-4, $d$:0.4, $p$:32, $\epsilon$:0.8, $m$:100, $c$:20, $K$:1, $q$:16, $seed$ = 42\\

\bottomrule
\end{tabular}
\end{center}
\end{table*}

\section{Conclusion}

In this paper, we propose an effective GNN approach, referred to as GCN-SL.
This paper includes three main contributions, compared with other GNNs. 1) SC is integrated into GNNs for capturing long range dependencies on graphs, and we propose a ESC-ANCH algorithm for dealing with large-scale graph-structured data, efficiently; 2) From the aspect of edges, the proposed GCN-SL can learn a high quality re-connected adjacency matrix by using similarity learning and special data preprocessing technique so as to improve classification performance of GCN-SL; 3) GCN-SL is appropriate for all levels of homophily by combining multiple key node intermediate representations. Experimental results have illustrated the proposed GCN-SL is superior to other existing counterparts.
In the future, we plan to explore effective models to
handle more challenging scenarios where both graph adjacency matrix and node features are noisy.

\nocite{langley00}

\bibliography{GCN-EF}
\bibliographystyle{icml2021}

\end{document}